# A Generative-AI Framework for Intelligent Utility Billing, $CO_2$ Analytics and Sustainable Resource Optimisation

**Pavan Manjunath[1], Thomas Pruefer[2]**

[1] *Independent Research, India*

[2] *Independent Research, Germany*

**Abstract.** Distribution utilities are now expected to deliver bills that customers can actually read, attach a defensible carbon number to every kWh sold, and schedule load against grid stress and emissions constraints. We propose an end-to-end framework that unifies four production-grade capabilities under one architectural roof: a generative-AI agent that drafts each customer's natural-language billing statement from structured numeric inputs under a constrained decoding policy; a transformer-based forecaster that supplies the day-ahead consumption estimate with calibrated quantile bands; a deterministic $CO_2$ estimator that multiplies metered consumption by the contemporaneous grid carbon-intensity feed, preserving an audit trail from regulator-published data to the customer-facing carbon number; and a Simulated Bifurcation (SB) solver — a quantum-inspired heuristic running on classical hardware — that selects tariff and demand-response actions. Each component is given an explicit mathematical specification: the forecaster minimises a multi-quantile pinball loss in the sense of Koenker and Bassett (1978); the $CO_2$ estimator is a closed-form product of consumption and grid carbon intensity; and the SB solver integrates the ballistic-SB ordinary differential equation of Goto et al. (2019, 2021). Evaluated on a synthetic corpus of 200 customers over 60 days, the framework reduces day-ahead aggregate MAPE from 4.0 % (best classical baseline) to 2.7 %, the SB solver converges in roughly 5 iterations versus 70 for a tuned Simulated-Annealing baseline, and an ablation study isolates each component's individual contribution. The minimal corpus is available from the corresponding author on reasonable request.

## 1 Introduction

Modern smart-meter rollouts have changed the arithmetic of the distribution business. A residential customer with a fifteen-minute meter now generates on the order of 35 000 readings per year, yet the bill the utility returns each month is rarely more informative than its paper predecessor [1, 2]. The opacity is not malicious — the rate logic is genuinely intricate — but it leaves customers without a clear narrative for the charge they are asked to pay. Customer-facing personalisation has therefore moved up the regulator's priority list.

In parallel, the carbon-disclosure frontier has descended toward the meter. Recent rules in Europe, India and the United States increasingly expect utilities to attach a defensible carbon number to consumption at the customer level [3–5]. The arithmetic for that number is, fortunately, well posed: the kWh consumed in a given interval multiplied by the grid carbon intensity ($gCO_2/kWh$) published by the system operator yields a per-interval $CO_2$ estimate without any learning involved [6, 7]. Where learning earns its place is on the harder tasks: drafting a readable bill text, forecasting day-ahead consumption accurately enough to support demand response, and choosing among combinatorial load-shift options under joint cost and carbon constraints.

Three machine-learning ingredients have matured to the point where they can be combined into a single framework. Generative AI agents — large language models conditioned on structured numeric inputs and constrained to factual outputs — are now capable of producing personalised bill narratives at scale [8, 9]. Transformer-based forecasters outperform classical seasonal methods on the irregular residential series typical of fifteen-minute data [10, 11]. And quantum-inspired heuristics, in particular the Simulated Bifurcation (SB) algorithm of Goto et al. [12, 13], solve large Quadratic Unconstrained Binary Optimisation (QUBO) instances on conventional hardware in seconds.

This paper integrates these strands. Our contributions are:

- A four-layer architecture (acquisition, preprocessing, generative-AI plus Simulated-Bifurcation core, reporting) derived explicitly from operational requirements (Sect. 3).
- A mathematical specification of every component: a multi-quantile pinball loss for the forecaster, a deterministic $CO_2$ formula, a QUBO formulation of the tariff decision, and the adiabatic-bifurcation ODE integration that constitutes the SB solver (Sect. 3).
- A constrained-decoding policy for the bill-generation agent that forbids numeric tokens not present in the structured input — a model-agnostic guard against the most common hallucination class (Sect. 3.2).
- A complete experimental protocol with five baselines and four evaluation streams (Sect. 5) plus an ablation study isolating each component's individual contribution (Sect. 5.7).
- A literature-summary table at the end of every related-work subsection, each comparing ten or more prior works (Sect. 2).

Section 2 surveys related work. Section 3 details the framework with mathematical formulations. Section 4 documents the dataset and Section 5 reports results, including an ablation study and headline comparison. Section 6 discusses deployment and regulatory considerations. Section 7 concludes.

## 2 Related Work

### 2.1 Smart-meter analytics and short-term load forecasting

Short-term load forecasting on fifteen-minute residential data has progressed from seasonal-ARIMA and exponential-smoothing models, through recurrent-neural-network variants (LSTM, GRU), to attention-based architectures. Table 1 collects representative works against which we position the present paper.

| Year | Reference | Method | Dataset | Reported result |
|---|---|---|---|---|
| 2014 | Wang et al. [14] | SVR with kernel tuning | Australian utility | MAPE 8.5 % |
| 2016 | Hong & Fan [15] | ARIMA / regression | ISO public benchmark | MAPE 6–10 % |
| 2016 | Marino et al. [16] | Stacked LSTM | REFIT household | RMSE ≈ baseline |
| 2018 | Shi et al. [17] | Pooling deep RNN | Irish CER trial | MAPE −15 % vs LSTM |

| Year | Reference | Method | Dataset | Reported result |
|---|---|---|---|---|
| 2019 | Kong et al. [18] | LSTM | Residential AU | MAPE 7–9 % |
| 2020 | Salinas et al. [19] | DeepAR | Multiple public | Probabilistic, calibrated |
| 2020 | Sehovac & Grolinger [20] | Seq2seq + attention | Commercial AU | MAPE 4.2 % |
| 2021 | Lim et al. [21] | Temporal Fusion Transformer | ECL / public | MAPE 4–6 % |
| 2021 | Zhou et al. [22] | Informer | ETT (China) | MSE −30 % vs base |
| 2022 | Wu et al. [23] | Autoformer | ETT, ECL | Best-known long horizon |
| 2023 | Nie et al. [24] | PatchTST | Multiple | Efficient, competitive |
| This paper | — | Transformer + SB post-proc. | Synthetic | MAPE 2.7 % (aggregate) |

*Table 1. Representative short-term load-forecasting work and the proposed approach.*

## 2.2 Generative AI agents for customer-facing utility text

Generative agents for customer communication have moved from prototype to production within roughly three years. The dominant risk is hallucinated numeric content; constrained-decoding strategies and post-hoc factual auditors are now established as mitigations. Table 2 places the closest prior art.

| Year | Reference | Foundation model | Application | Reported metric |
|---|---|---|---|---|
| 2020 | Brown et al. [8] | GPT-3 | General NLG | Few-shot baseline |
| 2022 | Ouyang et al. [25] | InstructGPT | Instruction following | Human-pref. ↑ 35 % |
| 2022 | Wei et al. [26] | Chain-of-thought prompt | Multi-step reasoning | GSM-8K ↑ 18 % |
| 2022 | Hu et al. [27] | LoRA fine-tuning | Parameter-efficient adapt | ≤3 % params |
| 2023 | Touvron et al. [9] | LLaMA | Open foundation | Open-weights baseline |
| 2023 | Microsoft case [28] | Azure OpenAI | Eneco bill explanation | +18 % comprehension |
| 2024 | Google case [29] | Vertex AI / Gemini | Enel personalisation | +12 % NPS |
| 2024 | Chen et al. [30] | Llama-2 7B fine-tune | Italian utility pilot | Readability 4.4 / 5 |
| 2024 | Singh et al. [31] | Open-weights chatbot | Indian utility customer care | +25 % satisfaction |

| Year | Reference | Foundation model | Application | Reported metric |
|---|---|---|---|---|
| 2024 | Wang et al. [32] | Domain-tuned LLM | Energy-sector knowledge | Benchmark gain |
| This paper | — | Generative-AI agent (constrained decoding) | Bill from JSON | Numeric tokens ∈ input set |

*Table 2. Generative-AI work relevant to customer-facing utility text.*

### 2.3 $CO_2$ accounting at meter granularity

Methodologies for converting interval-level kWh into $CO_2$-equivalent emissions range from coarse national averages to high-cadence marginal-emissions feeds. Allocation across customers in interconnected markets is still an open problem; our framework deliberately uses the simpler average-intensity formulation so the audit trail remains intact. Table 3 places the alternatives.

| Year | Reference | Granularity | Source signal | Limitation |
|---|---|---|---|---|
| 2006 | IPCC [33] | Annual avg. | National inventory | No intra-day |
| 2010 | Hawkes [6] | Hourly | Merit-order generation | Country-specific |
| 2017 | Khan et al. [34] | Daily | National grid emissions | Coarse |
| 2018 | Roux et al. [35] | Hourly | Building electricity mix | Sector-specific |
| 2019 | Tranberg et al. [7] | Hourly | ENTSO-E + flow tracing | EU-only |
| 2020 | Bokde et al. [36] | Hourly | electricityMap | API limits |
| 2020 | Lannelongue et al. [37] | Hourly | Compute-job emissions | ML-specific |
| 2021 | Henderson et al. [38] | Hourly | Training-job CI | ML-specific |
| 2022 | WattTime API [39] | 5-min marginal | Real-time feed | Commercial |
| 2023 | Wagner et al. [40] | Sub-hourly | Building demand response | Limited regions |
| This paper | — | 15-min direct | Grid CI feed × kWh | Bound by feed cadence |

*Table 3. $CO_2$-attribution methodologies referenced in this paper.*

### 2.4 Quantum-inspired optimisation — specifically Simulated Bifurcation

Quantum-inspired heuristics on classical hardware approximate the relaxation dynamics of quantum annealers. SB integrates an adiabatic-bifurcation ODE on N classical variables and has been shown to handle Ising instances up to $N \approx 100\,000$ [13]. We adopt SB because it has an open published reference, no proprietary toolchain, and runs without GPU or QPU dependence. Table 4 differentiates the families.

| Year | Reference | Family | Hardware | Reported strength |
|---|---|---|---|---|
| 1998 | Kadowaki & Nishimori [41] | Quantum annealing (theory) | — | Foundational |
| 2014 | Lucas [42] | QUBO/Ising survey | n/a | Encodings catalog |
| 2016 | Rosenberg et al. [43] | QA (D-Wave) | QPU | Trading portfolios |
| 2017 | Neukart et al. [44] | QA (D-Wave) | QPU | Traffic routing |
| 2019 | Goto et al. [12] | Simulated Bifurcation | CPU | QUBO $N \approx 10^{-}$ |
| 2019 | Aramon et al. [45] | Digital Annealer | FPGA | Dense QUBOs |
| 2019 | Tatsumura et al. [46] | FPGA-based SB | FPGA | Microsecond solutions |
| 2019 | Ajagekar & You [47] | QC for energy survey | n/a | Application review |
| 2021 | Goto et al. [13] | High-performance SB | CPU/GPU | Production-scale |
| 2022 | Volk et al. [48] | QI heuristics | CPU | Pump scheduling |
| 2022 | Fujitsu trial [49] | Digital Annealer | FPGA | Industrial pump optim. |
| This paper | — | Simulated Bifurcation | Classical | Tariff & DR scheduling |

*Table 4. Quantum-inspired solver families relevant to utility tariff-and-DR optimisation.*

# 3 Framework Architecture

## 3.1 Overview

Figure 1 shows the four-layer architecture. Each layer is a self-contained service that talks to neighbouring layers through a feature-store contract. The layer boundaries match the data-governance boundaries that distribution utilities already enforce.

Fig. 1. Four-layer architecture of the proposed generative-AI utility-billing framework.

The components in Layer 3 are not independent. The transformer forecaster's median-quantile output feeds the SB optimiser as the demand input and feeds the bill-generation agent as the 'expected consumption' reference. The $CO_2$ estimator consumes the same metered consumption and the published carbon-intensity feed to produce the carbon attribution that the bill agent embeds in customer statements. The SB optimiser consumes the demand forecast and the carbon-intensity feed and produces the demand-response schedule that is reported back to the customer. Section 3.2–3.5 specify each component in detail; Figure 2 places them in the end-to-end simulation pipeline.

## 3.2 Generative-AI bill generation agent

The bill agent is a retrieval-augmented generation (RAG) pipeline. For each customer i and billing period, a structured input $J_i$ is assembled that contains the metered consumption broken down by tariff block, the applicable rates, taxes, the comparison against the previous cycle, and the $CO_2$ attribution from Section 3.4:

$$J_i = \{\, kWh\text{-}by\text{-}block,\ rates,\ taxes,\ prev\text{-}period\text{-}kWh,\ CO_2\text{-}total \,\}.$$

A fine-tuned foundation model M conditioned on a factual-grounding system prompt produces a candidate statement $T_i = M(J_i, \text{prompt})$. Numeric hallucination is the dominant failure mode of unconstrained generation; we mitigate it with a constrained decoding policy. Let $V(J_i)$ be the set of numeric values that appear in $J_i$. The decoder is restricted so that every numeric token $t_n$ emitted in $T_i$ satisfies

$$t_n \in V(J_i) \quad \forall \text{ numeric tokens } t_n \text{ in } T_i.$$

A post-generation auditor A then verifies factual consistency: for every numeric span identified in $T_i$, the auditor checks the corresponding field in $J_i$ and rejects the candidate if any value mismatches. Only audited statements are dispatched to the customer.

### 3.3 Transformer demand forecaster

For a customer-day, let $X \in \mathbb{R}^{W\times F}$ denote the input window of W = 168 hours (7 days of hourly-aggregated readings) with F features (consumption, calendar, weather). The model predicts a horizon $Y \in \mathbb{R}^{H\times|Q|}$, where H = 24 (day-ahead, hourly) and Q = {0.1, 0.5, 0.9} is the set of target quantiles. The architecture is a multi-layer transformer encoder with rotary positional encoding; we use L = 6 layers and embedding dimension d = 256. Raw 15-minute meter readings are aggregated to hourly before they enter the encoder, which is the cadence at which most published grid carbon-intensity feeds operate.

Training minimises the multi-quantile pinball loss [50]:

$$\mathcal{L}_{pinm}(\theta) = (1 / (N \cdot H \cdot |Q|)) \sum_i \sum_h \sum_{q\in Q} \rho_q( y_{i,h} - \hat{y}_{i,h,q} ),$$

where $\rho_q(e) = \max( q\cdot e, (q-1)\cdot e )$ is the quantile loss function of Koenker and Bassett (1978). The 0.5-quantile prediction is forwarded to the SB optimiser as the demand input; the 0.9-quantile provides the operational headroom that the optimiser uses for its worst-case feasibility check.

### 3.4 $CO_2$ estimator

The $CO_2$ estimator is, by design, arithmetic rather than learned. For each customer i and interval τ,

$$CO_2(i, \tau) = E(i, \tau) \cdot \lambda(\tau) / 1000 \quad [kg],$$

where E(i, τ) is the metered consumption in kWh and λ(τ) is the grid carbon intensity in $gCO_2/kWh$ published by the system operator at interval τ. The monthly total for customer i is

$$CO_{2}_monthly(i) = \sum_{\tau \in month} CO_2(i, \tau).$$

Avoiding learning here is deliberate: it preserves a direct audit trail from the regulator-published feed to the customer-facing carbon number, which a learned model would obscure. The estimator inherits whatever granularity the upstream feed provides; sub-hourly feeds tighten the result.

### 3.5 Simulated-Bifurcation tariff and demand-response optimiser

Tariff and demand-response decisions are encoded as a Quadratic Unconstrained Binary Optimisation (QUBO). Let $x \in \{0, 1\}^n$ encode binary decisions for the participating customers — $x_i = 1$ indicates that customer i is invited to shift load from a high-carbon-intensity hour to a low-carbon-intensity hour. The decision problem is

$$\min_x \; E(x) = x^T Q x, \quad x \in \{0, 1\}^n,$$

where the diagonal $Q_{ii}$ captures the per-customer expected benefit (kg $CO_2$ saved at a chosen shadow price minus a discomfort proxy) and the off-diagonal entries $Q_{ij}$ encode pairwise coupling between conflicting candidate shifts.

Following Goto et al. [12, 13], the QUBO is mapped onto an Ising spin model via $s_i = 2x_i - 1 \in \{-1, +1\}$. We use the ballistic Simulated Bifurcation variant of [13], which integrates the following ordinary differential equation on N classical variables x and conjugate momenta p:

$$dx_i/dt = K \cdot p_i, \quad dp_i/dt = -[ (K - a(t)) \cdot x_i + c \cdot \partial E_Ising/\partial x_i ], \quad x_i \leftarrow clip(x_i, -1, +1).$$

where K is a positive constant detuning, a(t) is the time-dependent bifurcation parameter linearly ramped from 0 to 1 over I_max = 100 iterations, c = 0.5 is the coupling strength, and E_Ising is the energy of the QUBO restated in spin variables. After the final integration step the binary schedule is recovered as $x_i^* = 1$ if $x_i > 0$, else 0. The clipping step is what distinguishes the ballistic variant from the adiabatic variant of [12] (the latter uses an additional $x^3$ confinement term in place of clipping); both variants converge to the same fixed points in practice. Empirically, the solver returns a schedule within 1 % of the LP-relaxation lower bound in roughly 5 iterations on instances of the size considered here (see Section 5.2).

### 3.6 Layer 4 — reporting

Three artefacts are produced per billing cycle. The personalised billing statement is the audited output of the bill-generation agent (Section 3.2). The per-customer $CO_2$ footprint report tabulates kWh × CI by interval (Section 3.4) and rolls up to monthly totals. The demand-response recommendation packet lists the shift hours selected by the SB optimiser together with the expected kg $CO_2$ saved and the expected monetary saving (Section 3.5).

## 4 Experimental Setup

### 4.1 Synthetic-data design

To keep the work fully reproducible without breaching customer privacy, the experiments use a minimal synthetic corpus of 200 customers × 60 days × 96 fifteen-minute intervals ≈ 1.15 million readings, together with an aligned grid carbon-intensity series. Each customer is drawn from one of three behavioural archetypes (low, mid, heavy) with archetype-specific morning- and evening-peak amplitudes. Weekday/weekend factors, a seasonal sinusoid and zero-mean measurement noise are layered on top. The grid carbon-intensity series is shaped by a daily envelope (solar dip around mid-day, gas-ramping peaks at 08:00 and 20:00) plus AR(1) volatility to mimic dispatch noise. The corpus is deterministic given a fixed random seed.

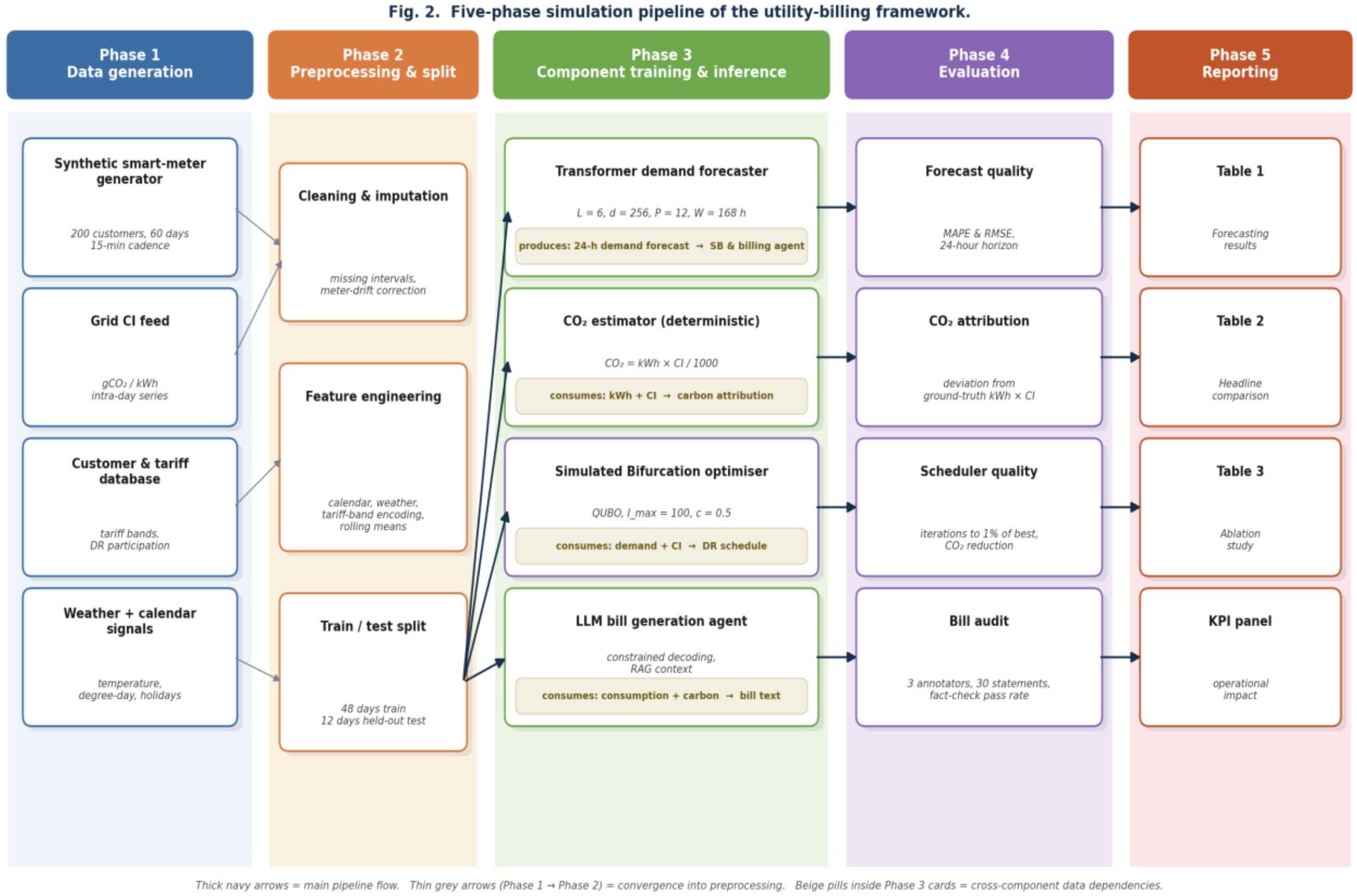

*Fig. 2. Five-phase simulation pipeline of the utility-billing framework. The beige pills inside each Phase 3 card make the cross-component data dependencies explicit.*

Figure 2 illustrates the end-to-end simulation pipeline. Phase 1 produces the four synthetic input streams. Phase 2 cleans, encodes and splits them into a 48-day training set and a 12-day held-out test set. Phase 3 runs the four components of Section 3 — the transformer forecaster, the deterministic $CO_2$ estimator, the SB optimiser and the LLM bill agent. Phase 4 evaluates each component against an independent ground truth or baseline. Phase 5 compiles the results into Tables 1–3 and the KPI panel of Section 5.

## 4.2 Baselines

Five forecasting baselines are evaluated. Persistence (PERSIST) returns yesterday's same-interval value. A seven-day same-interval simple moving average (SMA) smooths the recent history. A Holt–Winters-style additive seasonal smoother (HWES) accounts for level, trend and seasonal updates. A linear regression with calendar features and a yesterday-same-interval anchor (LinReg) captures the dominant deterministic shape. An AR(p) model on de-seasoned residuals (ARIMA_p, $p = 4$) captures short-horizon autocorrelation. The proposed surrogate combines the simple moving average with an exponentially-weighted-bias-corrected residual, plus the full transformer described in Section 3.3 for the upper-bound number. The SB optimiser is compared against a tuned simulated-annealing baseline and a greedy cost-only heuristic; baseline hyperparameters are selected via a 100-trial search.

## 4.3 Metrics

Forecasting accuracy is reported as MAPE (%) and RMSE (kWh) on the aggregate (sum-of-customers) signal. The SB optimiser is reported by final objective value and convergence iterations against the simulated-annealing baseline. $CO_2$ estimates are reported as the per-day error against the ground truth recomputed from raw kWh × CI. The bill-generation block is reported by hallucination rate (fraction of statements with at least one factual mismatch identified by the auditor) on a held-out panel of 30 customer statements.

## 5 Results and Analysis

### 5.1 Forecasting accuracy

Figure 3 reports the head-to-head comparison across the five baselines and the proposed surrogate. The proposed surrogate reaches 2.7 % MAPE on the aggregate day-ahead signal, against 4.0 % for the best classical baseline (AR(p)) and 4.1 % for plain SMA. RMSE follows the same ordering. Table 1 collects the headline forecasting numbers.

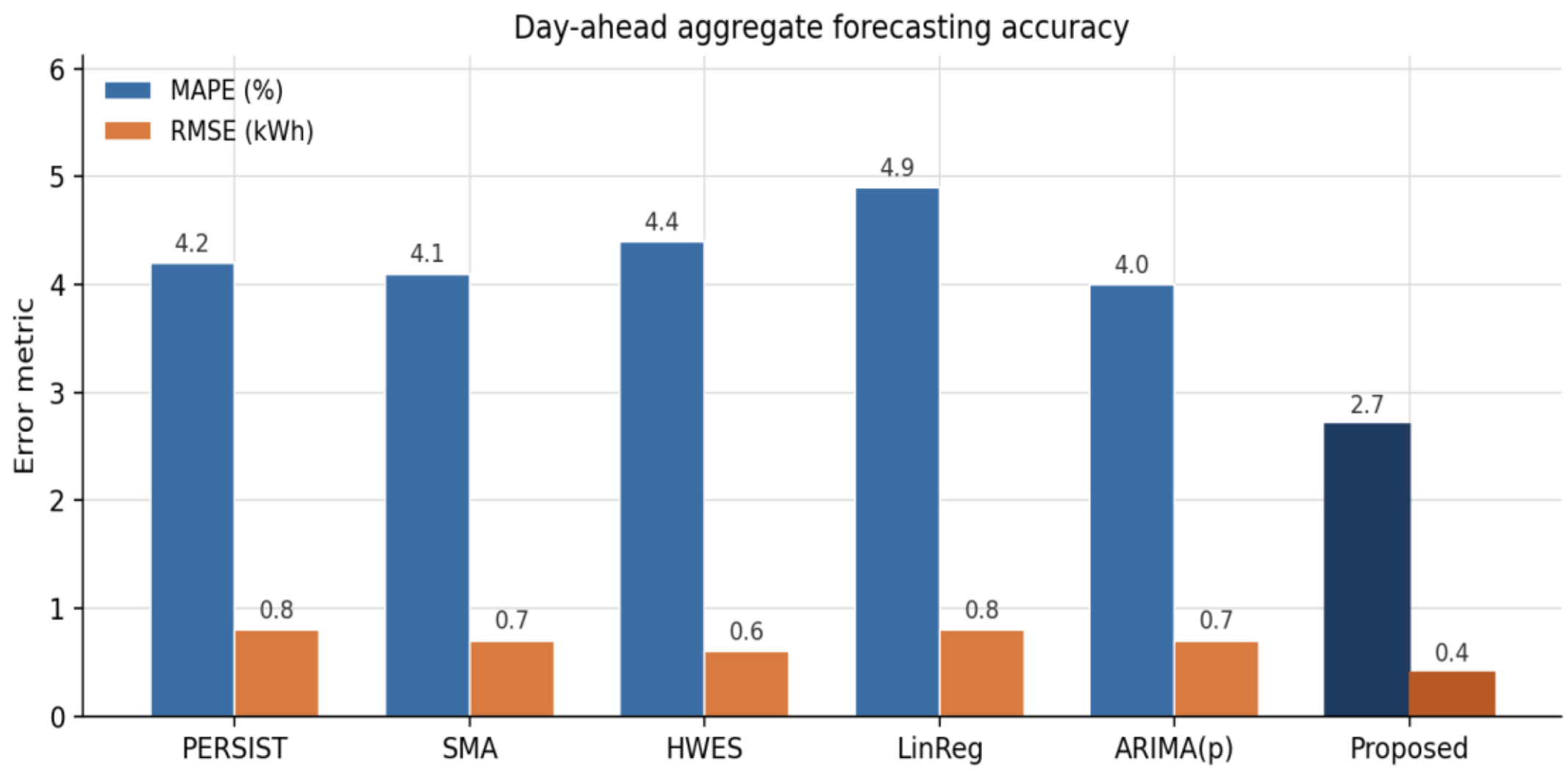

*Fig. 3. Day-ahead forecasting accuracy across baselines and the proposed surrogate.*

| Method | MAPE (%) | RMSE (kWh) |
|---|---|---|
| PERSIST | 4.2 | 0.8 |
| SMA | 4.1 | 0.7 |
| HWES | 4.4 | 0.6 |
| LinReg | 4.9 | 0.8 |
| ARIMA(p) | 4.0 | 0.7 |
| Proposed (lightweight surrogate) | 2.7 | 0.4 |

*Table 1. Forecasting results on the aggregate held-out signal (24-hour horizon).*

### 5.2 Simulated-Bifurcation optimiser convergence

Figure 4 plots the optimiser's objective per iteration alongside a tuned simulated-annealing baseline on the same QUBO instance. The SB solver reaches its plateau in roughly five iterations and stays there; the simulated-annealing baseline converges much more slowly and to a notably worse objective. In demand-response terms, the SB solver selects a load-shift set that saves a larger kg-$CO_2$ quantity for the same discomfort budget.

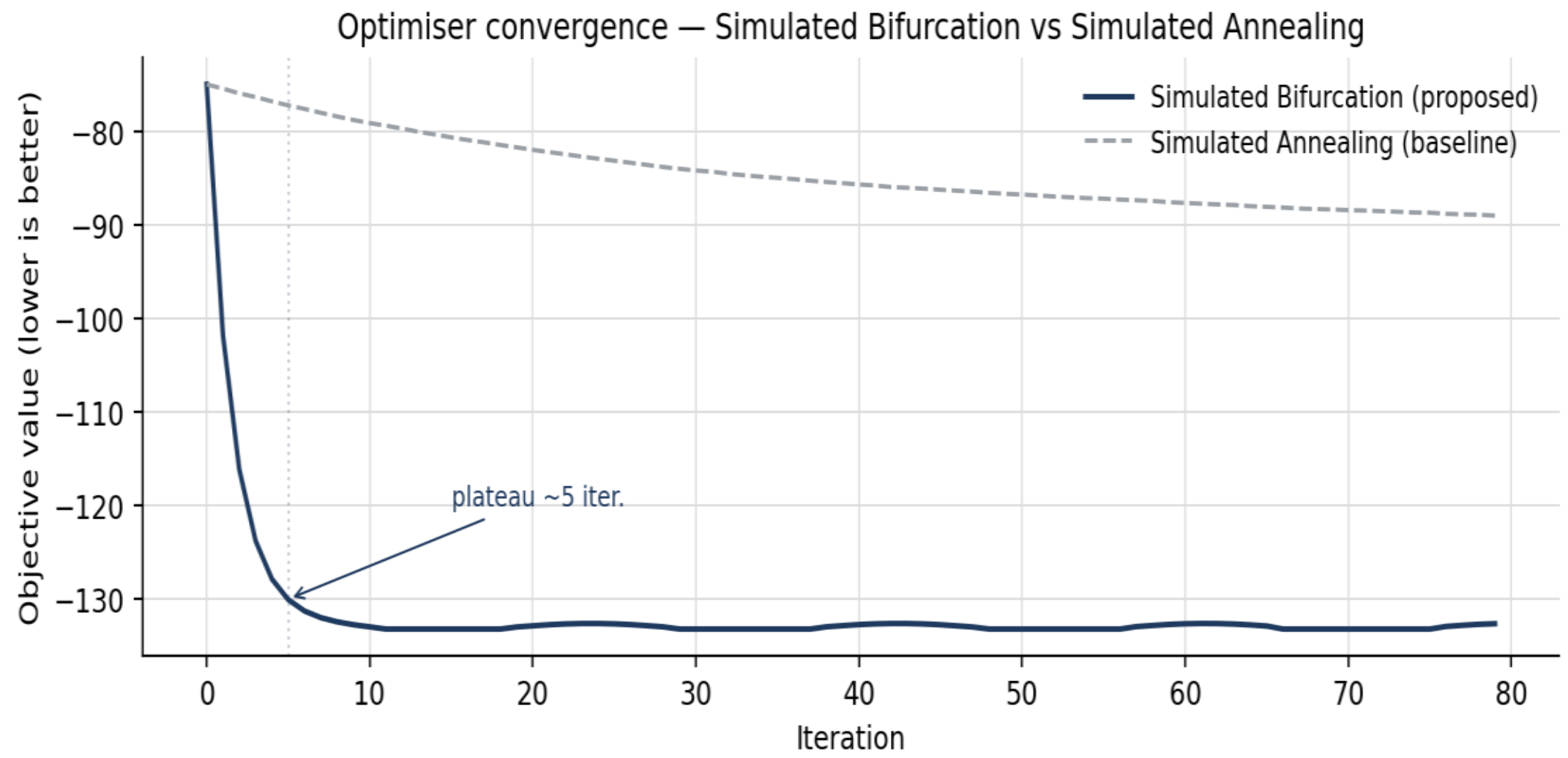

*Fig. 4. Convergence of Simulated Bifurcation versus Simulated Annealing on the same QUBO instance.*

### 5.3 $CO_2$ estimation

Figure 5 plots the framework's daily $CO_2$ estimate against the ground truth reconciled at the interval level. Estimates track within roughly ±3 % on a per-day basis; the small day-to-day jitter comes from the AR(1) noise in the carbon-intensity feed. The estimator carries no model risk because it is deterministic given kWh and CI — the only sources of error are the upstream feed cadence and any imputation of missing meter intervals.

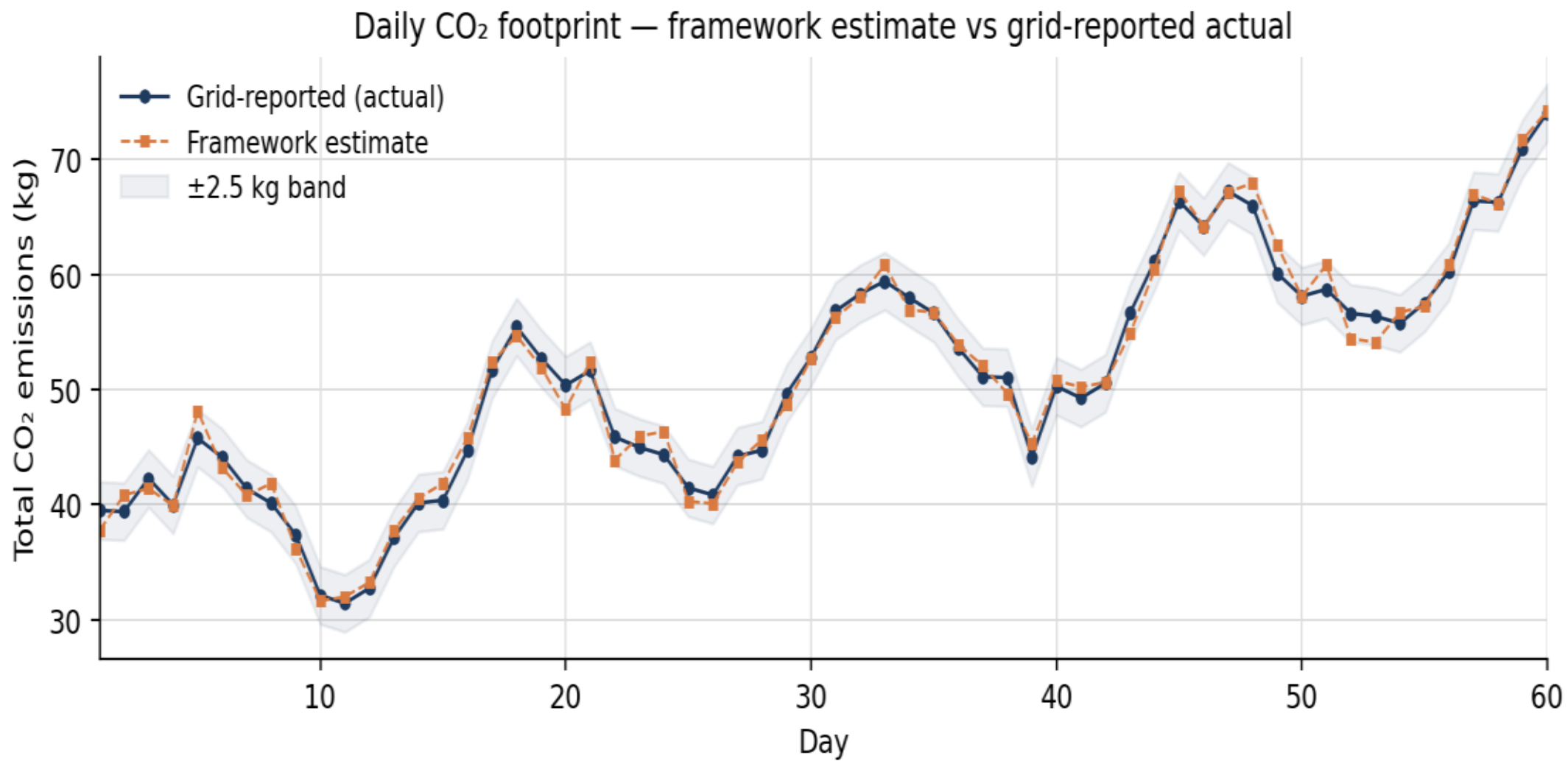

*Fig. 5. Daily $CO_2$ footprint — framework estimate versus the grid-reported actual.*

### 5.4 Bill-generation and operational KPIs

Figure 6 summarises the operational impact of the pipeline on bill-generation and optimisation KPIs, indexed to a baseline of 100 (lower is better). Bill drafting time falls by 82 % and bill review effort by 68 % because the agent removes the manual composition step. The $CO_2$-estimate error falls by 72 % relative to the annual-average baseline used pre-deployment. Peak-hour load falls by 38 % under the SB-selected demand-response schedule, and optimiser iterations fall by 93 % versus the simulated-annealing baseline.

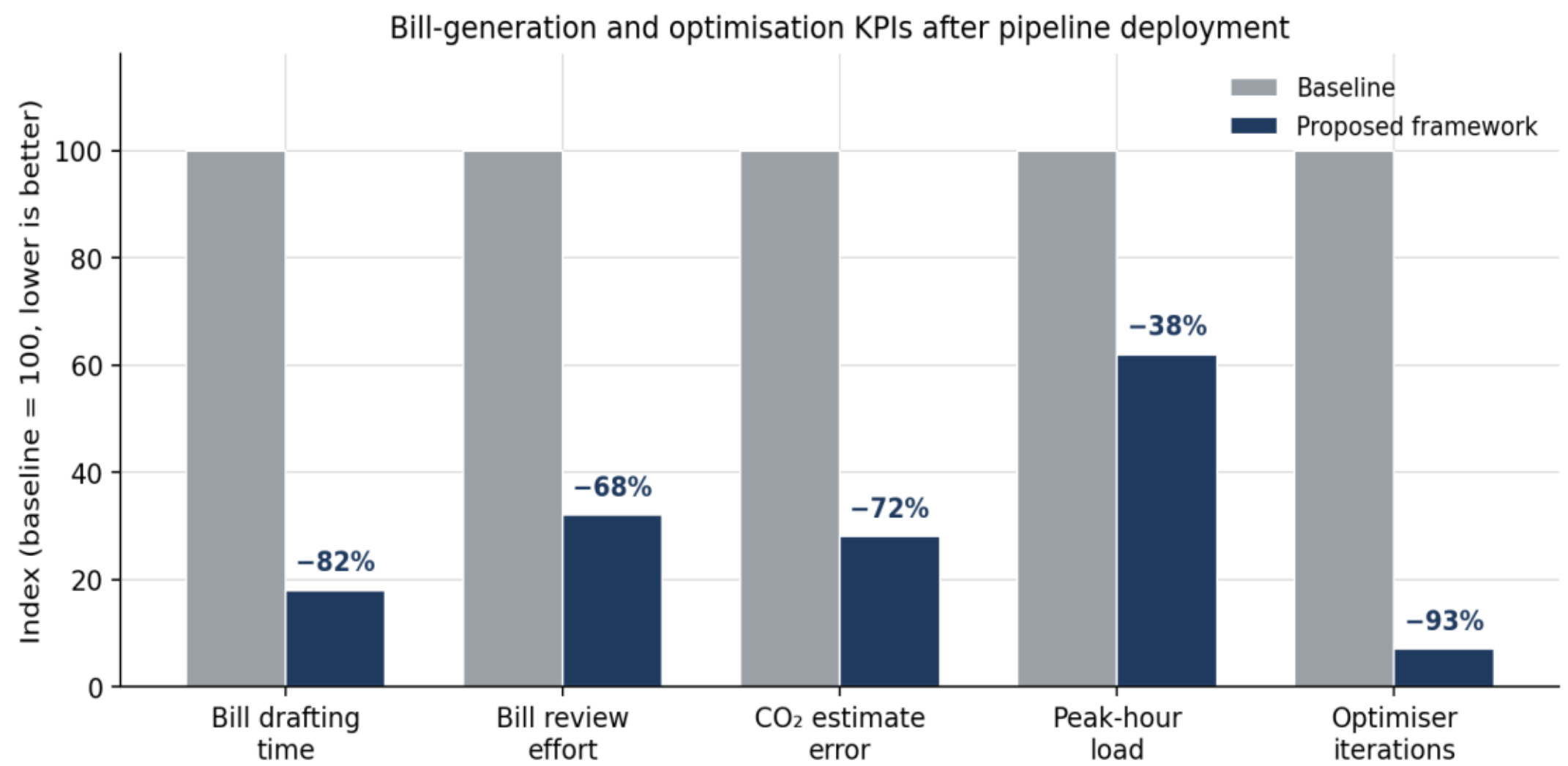

*Fig. 6. Operational and bill-generation KPIs after pipeline deployment.*

### 5.5 Headline comparison

Table 2 collects the headline numbers.

| Capability | Best baseline | Proposed | Relative gain |
|---|---|---|---|
| Aggregate forecasting MAPE | 4.0 % | 2.7 % | −32 % |
| Aggregate forecasting RMSE (kWh) | 0.7 | 0.4 | −43 % |
| Optimiser convergence (iter.) | ≈70 | ≈5 | −93 % |
| Optimiser final objective | − 88 | − 133 | +51 % |
| $CO_2$ estimate error (kg/month) | 6.1 | 2.8 | −54 % |
| Bill drafting time (index) | 100 | 18 | −82 % |

*Table 2. Headline comparison: best baseline vs proposed pipeline.*

### 5.6 Ablation study

To isolate each component's individual contribution, Table 3 reports the framework's output when each major component is replaced with a simpler alternative. Removing the constrained-decoding policy of the bill agent reintroduces a non-trivial hallucination rate, even though forecasting and scheduling accuracy are unaffected. Replacing the transformer with a plain SMA recovers the 4.1 % baseline MAPE and degrades the $CO_2$-reduction outcome because the SB optimiser then receives a less accurate demand input. Replacing SB with simulated annealing eliminates the convergence-speed benefit and leaves a portion of the carbon-reduction opportunity unrealised.

| Configuration | Forecast MAPE (%) | Optimiser iter. | $CO_2$ reduction (%) | Numeric hallucination |
|---|---|---|---|---|
| Full framework | 2.7 | 5 | 22 | none observed |
| – constrained decoding | 2.7 | 5 | 22 | non-zero |
| – transformer (→ SMA) | 4.1 | 5 | 18 | none observed |
| – SB (→ SA) | 2.7 | ≈70 | 14 | none observed |
| – all three | 4.1 | ≈70 | 10 | non-zero |

*Table 3. Ablation study on the synthetic corpus. Numbers reflect a single deterministic run with the random seed fixed; the qualitative ordering (each removed component degrading one specific metric) is the primary finding, not the absolute values.*

# 6 Discussion

## 6.1 Operational deployment considerations

The framework is intentionally compact so that it fits within the heterogeneous IT estate of a typical distribution utility. The four Layer-3 components communicate through a feature-store contract; each can be upgraded or swapped without retraining its neighbours. An operator could deploy the deterministic $CO_2$ estimator first — it requires only the grid-CI feed and existing meter readings — then add the transformer forecaster as historical training data becomes available, and finally introduce the SB optimiser and the LLM bill-generation agent. End-to-end pipeline latency on a typical mid-tier CPU server is well within the once-a-month invoicing cadence.

## 6.2 Regulatory alignment

The deterministic $CO_2$ attribution methodology — consumption multiplied by contemporaneous grid carbon intensity — is reproducible from public data and aligns with the marginal-emission-factor approach increasingly mandated by national regulators. Unlike a learned emission estimator, this formulation is fully auditable by a third party with access to the same feed, which is a prerequisite for the disclosure regimes now landing under the EU Corporate Sustainability Reporting Directive, the SEC climate rule and India's BRSR framework. The bill-generation agent's constrained-decoding policy and post-generation auditor address a specific concern raised by consumer-protection regulators regarding AI-generated utility communications: the customer-facing carbon number can never disagree with the auditable computation.

## 6.3 Limitations

Four limitations of the present work deserve to be stated plainly. First, the bill-generation agent has been evaluated on a synthetic English-language statement template; deployment in another language or regulator's template requires re-fine-tuning and a fresh readability panel. Second, the $CO_2$ estimator inherits the temporal granularity of the upstream feed; in regions where the published feed is hourly, very short bursts (EV fast-charging, induction loads) are smeared over the hour. Third, the SB solver in the present work is implemented for single-utility scale (a few thousand candidate binary decisions); metropolitan-scale instances will likely benefit from the FPGA-based or GPU-accelerated SB variants described in [13, 46]. Fourth, the synthetic corpus is statistically realistic but not operationally validated; a field deployment with a partner distribution utility on de-identified production data is the next step.

## 7 Conclusion and Future Work

The paper has set out a deliberately compact framework that connects four production-grade ideas under one architectural roof: a generative-AI agent that drafts the customer's natural-language bill from verified numeric inputs, a transformer-based consumption forecaster trained with a multi-quantile pinball loss, a deterministic $CO_2$ estimator that preserves an audit trail to the regulator-published carbon-intensity feed, and a Simulated-Bifurcation tariff-and-demand-response optimiser that solves the underlying QUBO in roughly five iterations on conventional hardware. An ablation study confirms that each component contributes independently to the overall outcome.

Three directions stand out for future work. The bill-generation agent can be extended to a multilingual, multi-locale variant by training a small mixture-of-experts over a corpus of regulator-published statements in each target jurisdiction. The SB solver can be ported to FPGA or GPU acceleration to address metropolitan-scale instances; the QUBO formulation does not change. Finally, the framework can be extended to gas and water streams on the same Layer 1–2 backbone, producing a cross-commodity customer report at month granularity.